\documentclass[a4paper,fleqn]{cas-dc}

\usepackage[numbers,sort&compress]{natbib}

\def\tsc#1{\csdef{#1}{\textsc{\lowercase{#1}}\xspace}}
\tsc{WGM}
\tsc{QE}

\usepackage{algorithm}
\usepackage{algorithmic}
\usepackage{multirow}
\usepackage{subfigure}
\usepackage{booktabs}
\usepackage[switch]{lineno} 
\usepackage{amssymb}
\usepackage{amsmath}
\usepackage{bm}
\usepackage{float}
\usepackage{graphicx}
\usepackage{color}
\usepackage{makecell}

\begin{document}
\let\WriteBookmarks\relax
\def\floatpagepagefraction{1}
\def\textpagefraction{.001}

\newcommand{\todo}[1]{{\color{black}#1}}
\newcommand{\err}[1]{{\color{black}#1}}

\shorttitle{Active Relation Discovery: Towards General and Label-aware Open Relation Extraction}    

\shortauthors{Li et al.}  

\title [mode = title]{Active Relation Discovery: Towards General and Label-aware Open Relation Extraction}  



%


\author[1]{Yangning Li}[
      orcid=0000-0002-1991-6698,
    ]



\ead{liyn20@mails.tsinghua.edu.cn}


\credit{Conceptualization of this study, Methodology, Experiments}

\affiliation[1]{organization={Shenzhen International Graduate School, Tsinghua University},
            city={Shenzhen},
            postcode={518055}, 
            state={Guangdong},
            country={China}}

\author[1]{Yinghui Li}[]
\credit{Conceptualization of this study, Methodology, Experiments}
\author[2]{Xi Chen}[]
\credit{Revision of the paper, Funding Support}
\author[1,4,]{Hai-Tao Zheng}[]
\cormark[1]
\credit{Revision of the paper, Funding Support}
\ead{zheng.haitao@sz.tsinghua.edu.cn}
\author[3]{Ying Shen}[]
\cormark[1]
\credit{Investigation process, Experimental verification}





\affiliation[2]{organization={Platform and Content Group, Tencent},
            city={Shenzhen},
            postcode={518055}, 
            state={Guangdong},
            country={China}}

\affiliation[3]{organization={school of Intelligent Systems Engineering, Sun Yat-Sen University},
            city={Guangzhou},
            postcode={510275}, 
            state={Guangdong},
            country={China}}

  \affiliation[4]{organization={Peng Cheng Laboratory},
            city={Shenzhen},
            postcode={518055}, 
            state={Guangdong},
            country={China}}
\cortext[1]{Corresponding author}



\begin{abstract}
Open Relation Extraction (OpenRE) aims to discover novel relations from open domains. Previous OpenRE methods mainly suffer from two problems: (1) Insufficient capacity to discriminate between known and novel relations. When extending conventional test settings to a more general setting where test data might also come from seen classes, existing approaches have a significant performance decline. (2) Secondary labeling must be performed before practical application. Existing methods cannot label human-readable and meaningful types for novel relations, which is urgently required by the downstream tasks. To address these issues, we propose the \textbf{A}ctive \textbf{R}elation \textbf{D}iscovery (\textbf{ARD}) framework, which utilizes relational outlier detection for discriminating known and novel relations and involves active learning for labeling novel relations. Extensive experiments on three real-world datasets show that ARD significantly outperforms previous state-of-the-art methods on both conventional and our proposed general OpenRE settings. The source code and datasets will be available for reproducibility.
\end{abstract}


\begin{highlights}
\item We reveal two major shortcomings of previous OpenRE: Insufficient capacity to discriminate between known and novel relations, and requiring additional secondary labeling. We also propose a practical test setup (General OpenRE) to appeal to the information extraction community to focus more on how the model performs in real-world scenarios.
\item We propose ARD, a framework that not only adapts to the General OpenRE utilizing relational outlier detection, but also exploits active learning to assign more meaningful and human-readable labels to novel relations. ARD offers a new, feasible and practical perspective for solving OpenRE.
\item Extensive experiments on both conventional and General OpenRE settings show that ARD achieve significant improvements in three real-world datasets.
\end{highlights}


\maketitle

\section{Introduction}\label{sec:introduction}
Novel relations are cropping up at a rate of tens of thousands per year~\cite{shi2018open}, while most of the rapidly emerging relations are still unlabeled and under-explored, mixed with pre-defined relations. These relations cannot be well handled by supervised RE methods due to the fixed pre-defined relation schema. Therefore, Open Relation Extraction (OpenRE) aims at discovering and extracting potential novel relations from open-domain corpora.

Some recent preliminary studies~\cite{wu2010open,wu2019open} have noticed the challenge of learning emerging relations and explored methods for OpenRE. Previous works can be divided into two main paradigms: pattern-based and clustering-based methods. Specifically, pattern-based methods~\cite{angeli2015leveraging,cui2018neural} utilize statistical or neural approaches to heuristically extract relation patterns from sentences, then clustering-based methods~\cite{elsahar2017unsupervised,wu2019open} are proposed to aggregate instances representing the same novel relation.
However, previous works mainly have two shortcomings in real scenarios:

(1) \textbf{The widely used traditional setting can't comprehensively reflect what OpenRE in the real world entails.} 
The traditional OpenRE setup is that models are evaluated based on their ability to discriminate among unseen classes, assuming the absence of known relation during the test phase. 
This test setup is a good measure of the model's ability to learn novel relations, but ignores the model's ability to distinguish between the known and unseen relations.
As we all know, the relation distribution in the real world is intricate, mixed with known and unseen relations. 
Therefore, it's unrealistic to assume that we will never encounter known relations during the test stage. 

In the light of above facts, we loosen the existing setting to a \textit{General OpenRE} setting: test data might also come from known relations. Empirical experiments in Table~\ref{table-main-ori} show that the previous state-of-the-art OpenRE models~\cite{wu2019open,hu2020selfore,zhang2021open} perform poorly under this setting.

(2) \textbf{The results produced by previous OpenRE models require secondary labeling before they can be practically applied.} In other words, for a certainly discovered novel relation, the model cannot assign it a surface name with a specific semantic meaning. As the foundation of a series of downstream tasks, labels with actual meaning are inevitable. However, due to the absence of human knowledge, both pattern-based and clustering-based OpenRE methods lack the ability to name novel relation types as human-readable and meaningful. Pattern-based methods rely heavily on the surface phrase, yet relations between entities are often not directly represented by the span in the sentence. Clustering-based methods merely cluster instances that express the same relations, but do not provide concrete representations of the novel relations. Both methods require manual re-labeling of the novel relations found. This gap between model and practice hinders model application in real-world scenarios.

\begin{figure*}
    \centering
    \includegraphics[width = 0.98 \linewidth]{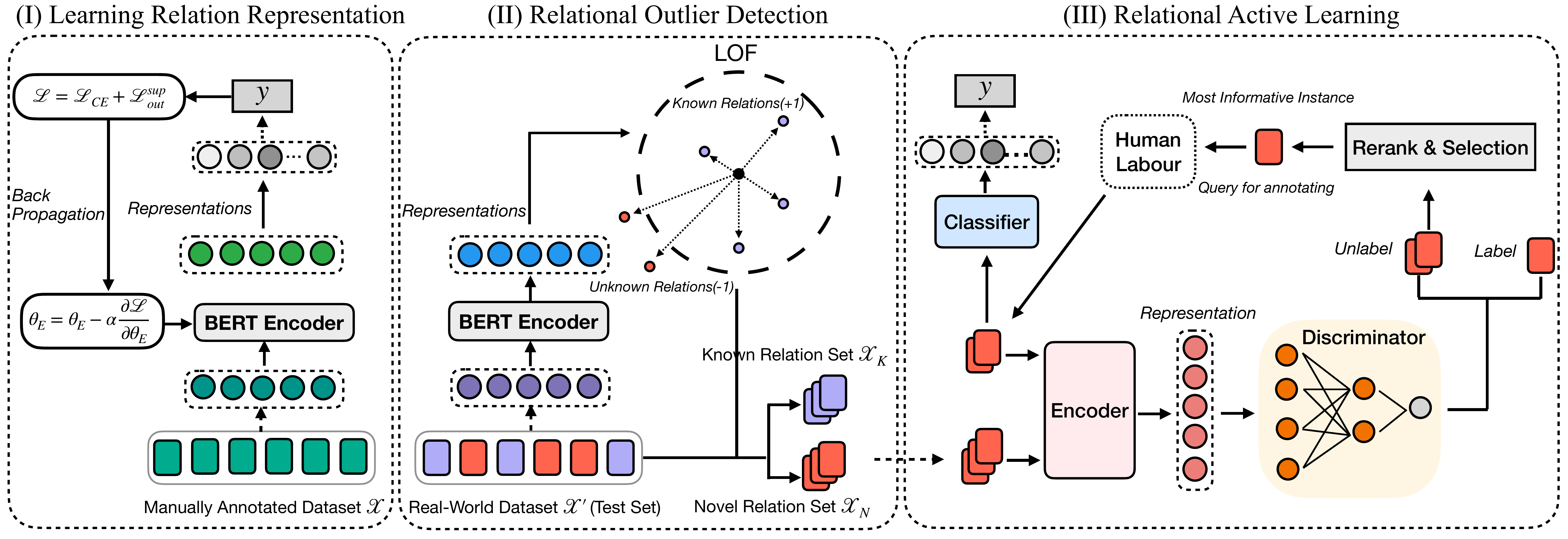}
    \caption{An illustration of our proposed Active Relation Discovery (ARD) framework.}
    \label{fig:model}
\end{figure*}

To address above mentioned issues, we propose the \textbf{Active Relation Discovery (ARD)} framework shown in Figure~\ref{fig:model}. Targeted improvements are made in two aspects: (1) To avoid the model being confused by the set of mixed known and novel relations, we developed a relational outlier detection algorithm that separates known and novel relations by treating novel relations as outliers, performing stably under the General OpenRE setting.
(2) To assign meaningful labels to novel relations, the incorporation of human knowledge is inevitable. To minimize the labor cost, we propose an active learning algorithm. Specifically, we introduce the \textit{representative instance}, which denotes an instance can offer rich information of unknown relations. Only a handful of representative instances requires manual labeling, and then the model can automatically label the novel relations in a supervised manner.

In summary, our contributions are in three folds:

(1) We reveal two major shortcomings of previous OpenRE, and introduce a new setting called \textit{General OpenRE}, which can realistically measure the capabilities of the OpenRE model.

(2) We propose ARD, a practical framework that not only adapts to the \textit{General OpenRE} utilizing relational outlier detection, but also exploits active learning to assign more meaningful and human-readable labels to novel relations.

(3) We conduct extensive experiments on both conventional and General OpenRE settings to show that our framework can achieve significant improvements in three real-world datasets. Detailed analyses demonstrate the effectiveness of each component of our proposed ARD.

\section{Related Work}

\noindent\textbf{Open Relation Extraction. }Whereas supervised RE~\cite{liu2013convolution, zhang2015relation,zhao2019improving,kishimoto2020adapting} relies heavily on manual annotation and the inherent inadequacy of predefined relation schema, OpenRE gains increasing attention. The method of OpenRE can be broadly divided into two categories: pattern-based and clustering-based. Pattern-based approaches extract relation patterns from textual corpora~\cite{banko2007open,fader2011identifying,stanovsky2016creating,stanovsky2018supervised}. These methods apply heuristic algorithms to describe relations between marked entities with relation patterns consisting of several key phrases in texts. Due to the ambiguity of relations obtained by the pattern-based methods, the focus of research in recent years has been primarily on clustering-based methods.

Clustering-based method~\cite{shinyama2006preemptive,elsahar2017unsupervised,wu2019open,hu2020selfore,zhang2021open} cluster instances in the feature space into novel relation types.~\cite{wu2019open} enhances unsupervised clustering-based methods by introducing Siamese Network to measure instance similarity. Considering the inherent connection between OpenRE and relation hierarchies, ~\cite{zhang2021open} proposes a framework to effectively integrate hierarchy information into relation representations for better novel relation extraction.

As described in Section~\ref{sec:introduction}, there are two main problems with the current OpenRE: (1) They focus only on the discrimination of novel relations, supposing that test sets only have novel relations. (2) The model output is not directly usable by downstream tasks. In response, we propose a General OpenRE setup and incorporate outlier detection and active learning into OpenRE.

\noindent\textbf{Active Learning in Relation Extraction. } 
The key idea behind active learning~\cite{settles2009active} is that the learning algorithm is allowed to ask for true meaningful labels of some selected unlabelled instances. Various criteria~\cite{zhang2012unified,fu2013efficient,qian2014bilingual} have been proposed to choose these instances on traditional supervised RE tasks. To our best knowledge, we firstly integrate active learning into OpenRE, enabling meaningful tags of the novel relation type with the addition of human knowledge.

\noindent\textbf{Generalized Zero-Shot Learning(GZSL). }The motivation for the General OpenRE setting is similar to that of the GZSL. Traditionally, ZSL approaches~\cite{romera2015embarrassingly,zhang2015zero} assume that only the unseen classes are present in the test set. ~\cite{chao2016empirical} first challenged this implausible setting and proposed the GZSL setting: test data might also come from seen classes. GZSL approaches~\cite{rahman2018unified,huang2019generative} focus on mitigating the strong bias caused by known classes and preventing novel classes from being categorized as one of the seen classes. While in our General OpenRE setting, we concentrate more on the distinction between known and novel classes.

\section{Task Formulation}

General OpenRE formulates the task slightly differently from traditional OpenRE setting. 
The original train set is a large-scale manually annotated corpus $\mathcal{X} = \{x_j^{r_i} | r_i \in \mathcal{R}_K\}$, where relations in $\mathcal{R}_K$ are pre-defined as ``known relations''. 
Obviously, we assume that there exists a relation set $\mathcal{R}_N$ that contains ``novel relations'' in another corpus without annotations. In the real-world scenario, we need to process the dataset whose instances express relations both in $\mathcal{R}_K$ and $\mathcal{R}_N$, distinguish known and novel relations, then label each instance. 

Under this fact, we first consider the \textit{novel relation discovery}, in which we solely focus on the mining of unseen relations. At this stage, we pre-train the model on $\mathcal{X}$ and obtain a trained encoder $E$. 
Then for a concrete dataset (test set) $\mathcal{X}' =\{ x_{j}^{r_{i}}, x_{j}^{\prime r_{i}^{\prime}} | r_i \in \mathcal{R}_K, r'_i \in \mathcal{R}_N \}$.
The model will unsupervisedly divide $\mathcal{X}'$ into a ``known relation set'' $\mathcal{X}_K$ and a ``novel relation set'' $\mathcal{X}_N$. 

$\mathcal{X}_K$ can be easily labeled for sufficient information obtained from $\mathcal{X}$. Secondly, we focus on the \textit{annotation of novel relations} $\mathcal{X}_N$. In this phase, we integrate the intuition of active learning by utilizing limited labor to facilitate the novel relation annotation performance. Our model queries a small set of informative samples in $\mathcal{X}_N$ for manual labeling and then trains a classifier to annotate novel relations. 

\section{Methodology}

\subsection{Overview}
The overview of the method is illustrated in Figure~\ref{fig:model}. We will detailedly introduce our work into three components: (1) \textbf{Relation representation}, in which we extend to transform semantic relations into low-dimension dense representations. (2) \textbf{Relational Outlier Detection}, where the model automatically detects a novel relation set from real-world datasets and feeds them into the active learning stage.
(3) \textbf{Relational Active Learning}, where the model selects the most informative instances to train a powerful classifier for novel relation.

\subsection{Relation Representation}

Given a dataset $X = \{x_1,...,x_n\}$, an instance $x$ is a word (token) sequence $\{w_1,w_2,...,w_n\}$ with two marked entities $e_h$ and $e_t$. We use triplets of relation facts $(e_h, r, e_t)$ to denote that there is a relation $r$ between the marked entity pair. And $x^r$ indicates an instance that expresses the relation $r$. Specifically, we define four special markers $\langle e_h \rangle$, $\langle /e_h \rangle$, $\langle e_t \rangle$, and $\langle /e_t \rangle$ to locate the head entity and the tail entity. We denote the indices of $\langle e_h \rangle$ and $\langle e_t \rangle$ as $\text{START(h)}$ and $\text{START(t)}$. An instance is represented as:
\begin{equation}
\begin{split}
        x=...,\langle e_h \rangle,w_{\text{START}(h)+1},..., w_{\text{END}(h)}, \langle /e_h  \rangle,..., \\ \langle e_t \rangle,w_{\text{START}(t)+1},..., w_{\text{END}(t)}, \langle /e_t  \rangle,...
\end{split}
\end{equation}
We use pre-trained language model (i.e. BERT~\cite{devlin-etal-2019-bert}) to encode each token $w_t$ to the corresponding representation $\bm{h}_t \in \mathbb{R}^d$, where $d$ is denotes the dimension of representation vectors. 

For an instance $x_i \in S$, we use the concatenation of representations of two start positions ($w_{\text{START}(h)}$ and $w_{\text{START}(t)}$) as the representation of the relation:
\begin{equation}
    \bm{h}_r(x_i) = [\bm{h}_{\text{START}(h)}, \bm{h}_{\text{START}(t)}],
\end{equation}
These extra tokens play a similar role like position embeddings in conventional RE tasks~\cite{zeng2015distant}. The relation representation $\bm{h}_r(x_i)$ will be utilized to predict the relation type $r$. 

As mentioned previously, $\mathcal{X}$ are used to fine-tune the pre-trained language model. Notably, along with the traditional cross-entropy loss, we integrate a supervised contrastive loss $\mathcal{L}_{\text {out }}^{\text {sup }}$ \footnote{Scalar temperature parameter $\tau$ is 0.1 as in ~\cite{khosla2020supervised}. We refer to ~\cite{khosla2020supervised} for more details.}:
\begin{equation}
\mathcal{L}_{\text{out}}^{sup}=  \sum_{i \in I} \frac{-1}{|P(i)|}\sum_{p \in P(i)} \log \frac{\exp \left(\boldsymbol{z}_{i} \cdot \boldsymbol{z}_{p} / \tau\right)}{\sum_{a \in A(i)} \exp \left(\boldsymbol{z}_{i} \cdot \boldsymbol{z}_{a} / \tau\right)},
\end{equation}

Here, $P(i) \equiv\left\{p \in B \backslash\left\{i\right\}: \tilde{\boldsymbol{y}}_{p}=\tilde{\boldsymbol{y}}_{i}\right\}$ is the set of indices of all positives in the mini-batch $B$ distinct from $i$. $\boldsymbol{z}_{i}=Proj(\bm{h}_r(x_i)) \in \mathcal{R}^{D_{P}}$, where $Proj$ is  a single linear layer outputs vector of size $D_{P}=128$. Contrastive loss \cite{khosla2020supervised,10.1145/3477495.3531954,li2022past}allows for tighter clustering of intra-class instances and a more dispersed distribution of inter-class instances. The essence behind the employment of contrastive loss is to gain relation representations that are more friendly to outlier detection and active learning. The performance of our relation representation on supervised RE can also be found in Section~\ref{appendix_section_supervisre}.

\subsection{Relational Outlier Detection}

After pre-training, $\texttt{E}_\theta$ could encode an instance $x$ into a dense vector $\bm{h}_r(x)$ as the relation representation. In the feature space, due to the similarity of the semantics, representations that express the same relation tend to densely gather (forming $n$ separate clusters) and ones that express different relations tend to disperse.  Figure~\ref{tsne} illustrated the distribution of different representations. Since the instances express unseen relations have not been pre-trained, in other words, the model has not seen the semantics, the instances are not projected near any clusters. We utilize this property to design local outlier factor (LOF) to reflect the local density of instances in the feature space.

Formally, given any two representations $\bm{h}_r(x_i), \bm{h}_r(x_j)$ of instances $x_i, x_j$, we denote $d(\bm{h}_r(x_i), \bm{h}_r(x_j))$ as the Euclidean distance between them. Then, we define $k$-th distance, denoted as $d_k(\bm{h}_r(x_i))$, to represent the distance from $\bm{h}_r(x_i)$ to the $k$-th nearest neighbour. The reach-ability distance between $\bm{h}_r(x_i)$ and $\bm{h}_r(x_j)$ is:

\begin{equation}
\begin{split}
   \text{rd}_k(\bm{h}_r(x_i), \bm{h}_r(x_j)) = \text{max}\{d_k(\bm{h}_r(x_j)), \\ d(\bm{h}_r(x_i), \bm{h}_r(x_j))\},
\end{split}
\end{equation}

We then compute the density to measure the average distance of reach-ability distance:

\begin{equation}
    \text{den}_k({h}_r(x_i)) = 1 / \frac{\sum_{{h}_r(x_j) \in N_k({h}_r(x_i))}  {\text{rd}_k({h}_r(x_i),{h}_r(x_j))}}{|N_k({h}_r(x_i))|},
\end{equation}
where $N_k(\bm{h}_r(x_i))$ denotes all the points whthin in $k$-th distance of $\bm{h}_r(x_i)$.

The computation of local outlier factor is:
\begin{equation}
    \text{LOF}_k(\bm{h}_r(x_i)) = \frac{\sum_{\bm{h}_r(x_j) \in N_k(\bm{h}_r(x_i))}  \frac{\text{den}_k(\bm{h}_r(x_j))}{\text{den}_k(\bm{h}_r(x_i)}}{|N_k(\bm{h}_r(x_i))|},
\end{equation}
where the larger $\text{LOF}$ is, the more likely $\bm{h}_r(x_i)$ is an outlier point, i.e, an instance that expresses a novel relation. Our model could unsupervisedly detect the instances with novel relations.

\subsection{Relational Active Learning}

To this end, the model could divide the real-world dataset into a ``known relation set" $\mathcal{X}_K$ and an ``novel relation set" $\mathcal{X}_N$. In view of the fact that $\mathcal{X}_K$ can be conveniently and precisely annotated, we focus on labeling meaningful types for discovered instances in $\mathcal{X}_N$ in this subsection. 

To retrieve human-readable labels and avoid subsequent secondary labeling, we need to incorporate human knowledge into the relation learning phase through active learning. Our primary goal is to find a small part of instances with the most information and artificially label them. Then we use the labeled data to train a classifier in a supervised manner. The problem of how to find instances with most information essentially is the problem of how to find the instances that are most likely to express ``novel relations''. Inspired by this, we propose the following Relation Active Learning module:

In the beginning, we randomly label a small part of data in $\mathcal{X}_N$. The labeled dataset is denoted as $\mathcal{X}_{L}$ and the rest of the unlabeled data is denoted as $\mathcal{X}_{U}$. We assume that all the instances $x$ are i.i.d 
according to a latent distribution $P(x)$. Correspondingly, their labels are distributed by the conditional distribution $P(y|x)$.

\noindent \textbf{Neural Encoder} We adopt a neural encoder to learn the distribution of $\mathcal{X}_L$ and $\mathcal{X}_U$ in the latent feature space. Our framework is independent of the choice of neural encoders, in this case, we adopt BERT~\cite{devlin-etal-2019-bert} as the encoder. The goal of the neural encoder is to encode $\mathcal{X}_L$ and $\mathcal{X}_U$ into the same feature space and try to fool a discriminator to correctly predict if the instance is ``representative''. The loss function of the encoder is:
\begin{equation}
\label{eq:enc}
\begin{split}
    \mathcal{L}_{e}  = &- \mathbb{E}_{x\sim P_{\mathcal{X}_L}}[\log (D_\psi(E_\theta(x)))] \\ &- \mathbb{E}_{x\sim P_{\mathcal{X}_U}}[\log(1 - D_\psi(E_\theta(x)))],
\end{split}
\end{equation}
\noindent \textbf{Discriminator} A binary classifier (or a discriminator): $\mathcal{X} \rightarrow \{-1, 1\}$ is adopted to select the most informative samples. We utilize adversarial training to leverage the information of both $\mathcal{X}_L$ and $\mathcal{X}_U$. 
The discriminator is adversarially trained to accurately distinguish if the instance expresses a novel relation.\footnote{ A single novel relation where it won't be picked to be labeled will eventually be labeled as a novel relation that has already been labeled, but this almost never happens.} The loss function is a flipped version of the encoder:
\begin{equation}
\label{eq:dis}
\begin{split}
    \mathcal{L}_{d}  = &- \mathbb{E}_{x\sim P_{\mathcal{X}_L}}[\log (1 - D_\psi(E_\theta(x)))] \\ &- \mathbb{E}_{x\sim P_{\mathcal{X}_U}}[\log( D_\psi(E_\theta(x)))]
\end{split}
\end{equation}
Naturally, we could jointly optimize the two objective functions by allocate two parameters: $\mathcal{L} = \lambda \mathcal{L}_e + \lambda' \mathcal{L}_d$.

\noindent \textbf{Active learning} At each training step, we select $k$ instances with the highest confidence of the discriminator as the most informative instances. Then the instances will be manually annotated and then used to train the classifier. In our experiments, as in most active learning efforts, we use the golden label of the instance as the annotation result. At this point, the discussion of annotations needs to be further developed. 
Considering the explosive growth of the number of relations, 
an annotating process that supports online and continual learning of novel relations needs to be designed. Thus, we propose a practical and easy-to-implement annotation procedure. At the start, 
for each selected instance $x_i$, the annotator only needs to judge if $x_i$ has the same relation class as any instances of $\mathcal{X}_L$.  $x_i$ will be indexed as a novel relation if it doesn't share the same relation with instances in $\mathcal{X}_L$, or labeled as one known relation. 
After the procedure, the labels of relations would be easy to design than before the active learning begins. This manner effectively ensures the ability to continual learning and online learning of our framework, expediently fitting the real situation.
Subsequently, $\mathcal{X}_L$ will be fed into a classifier, which is a one-layer MLP~\cite{liu2022we} with an output layer, optimized by cross-entropy objective function, denoted as $\mathcal{L}_c$ and parameterized by $\gamma$:
\begin{equation}
    \mathcal{L}_c = \sum_{i \in |\mathcal{X}_L|} - \log p(y_L^{(i)}|x_L^{i}, \gamma).
\end{equation}


\begin{algorithm}[]
\caption{Training for Active Learner, $\lambda, \lambda', k$ are hyper-parameters.}
\label{alg:active}
\begin{algorithmic}
	\STATE {\bfseries Input:} Labeled data $(\mathcal{X}_L, Y_L)$, unlabeled data $\mathcal{X}_U$, initialized encoder model with $\theta$, discriminator model with $\psi$, classifier with $\gamma$ 
	\WHILE {not converge}
	\item Sample mini-batches $(x_L, y_L)$ from $(\mathcal{X}_L, Y_L)$
	Sample mini-batches ${(x_U)}$ from $(\mathcal{X}_U)$
	\item Compute $\mathcal{L}_{e}$ by Eq.~\ref{eq:enc}
	\item Update $\theta$ w.r.t $\mathcal{L}_e$
	\item Compute $\mathcal{L}_{d}$ by Eq.~\ref{eq:dis}
	\item Update $\psi$ w.r.t $\mathcal{L}_d$
	\item Select $k$ most informative instances $\{x_1,...,x_k\}$ by the output of $d$
	\FOR {$i \leftarrow 1$ to $k$ do}
	\IF{$x_i$ has the same relation as $x^{r}_j \in \mathcal{X}_{L}$}
	\STATE Label $x_i$ with $r$ and append $x_i$ to $\mathcal{X}_{L}$
	\ELSE
	\STATE Label $x_i$ with a new index and append $x_i$ to $\mathcal{X}_{L}$
	\ENDIF
	\ENDFOR
	\item Update $\gamma$ w.r.t $\mathcal{L}_c$
	\ENDWHILE
\end{algorithmic}
\end{algorithm}

\section{Experiments\label{sec:exp}}
In this section, we verify the performance of the model on three large-scale OpenRE datasets and their variants, and at the same time, a series of auxiliary experiments are carried out to prove the effectiveness of the model. Finally, we give a detailed analysis of the efficacy of our ARD framework.
\subsection{Baseline}
To demonstrate the effectiveness of our ARD models, we compare
our models with three state-of-the-art models:
(1) \textbf{RSN-CV}~\cite{wu2019open} employs conditional entropy and virtual adversarial learning to train Siamese Network to measure instance similarity.
(2) \textbf{SelfORE}~\cite{hu2020selfore} utilizes self-training to iteratively learn relation representations and clusters with the weak signals provided by large pretrained language model.
(3) \textbf{OHRE}~\cite{zhang2021open} integrate hierarchy information into relation representations for better novel relation extraction. For a fair comparison, we substitute all the encoding models in the baseline models with $\rm BERT_{LARGE}$.
\subsection{Datasets and Setting}
\noindent\textbf{Datasets }Three datasets and their variants are used to evaluate our model: FewRel~\cite{han2018fewrel}, New York Times Freebase(NYT+FB)~\cite{marcheggiani2016discrete} and FewRel2.0~\cite{gao-etal-2019-fewrel}, the first two of which have been widely used in previous RE works~\cite{simon2019unsupervised,hu2020selfore,zhang2021open}. We follow the division of the datasets from previous works. 

FewRel is one of the largest RE dataset. As in the previous work, we use the original train set of FewRel. The dataset contains 80 relation categories and 700 instances of each relation category. Among them, 64 relations are divided into the training set and the remaining 16 relations are chosen as the test set.

NYT+FB dataset aligns entities from the New York Times corpus with Freebase triplets. Following the setting in~\cite{simon2019unsupervised}, we filter out sentence with non-binary relations and obtain 41,000 labeled sentences containing 262 relations. The training and test sets comprise 212 and 50 relations respectively. 

To verify the cross-domain capability of the model comprehensively, we also use FewRel2.0 dataset whose training and test sets are from completely different domains. As an advanced version of FewRel, FewRel2.0 incorporates knowledge transferring. The test set of FewRel2.0 contains data of 10 relations (100 samples for each relation) in the biomedicine field, and the training set is exactly the same as FewRel.  The statistics of the data set are shown in Table~\ref{table-statics}.

\noindent\textbf{Datasets Processing }As described above, in the original OpenRE setting, there are no overlapping relations in the training and test sets. The relations in the test set are all novel relations. To measure the performance of the model in our proposed \textit{General OpenRE} setting, we resample the original dataset and gain two variants: \textit{noisy} and \textit{imbalanced}. In the test sets of the two variants, there exist known relations with different distributions. In other words, the original dataset corresponds to the conventional setting and the noisy and imbalanced variants to the general setting.

To obtain the noisy variant, we randomly select 40\% samples from original training sets. Given that the number of samples for each novel relation is identical in FewRel and FewRel2.0, we further construct the imbalanced variant to explore the performance of the model in the presence of class imbalance. Specifically, we build on the noisy variant by randomly discarding a portion of the samples with different probabilities for each relation class in the test set, yielding class imbalance in test set. The discarding probabilities for different relations are shown in the Table~\ref{table:dataset}.

\begin{table}[]
\caption{Statistical results for the dataset. \#CLS represents the number of relation types and \#SUM stands for the number of samples. In the addition equation $x+y$ in the table, $x$ and $y$ are the statistics for the known and novel relations separately.}
\centering
\scalebox{1}{
\renewcommand\arraystretch{1}
\setlength\tabcolsep{4pt}
\begin{tabular}{ccllll}
\toprule
\multirow{2}{*}{\bf{Dataset}}   & \multirow{2}{*}{\bf{Setting}} & \multicolumn{2}{c}{\bf{Train}} & \multicolumn{2}{c}{\bf{Test}} \\ \cmidrule(r){3-4}\cmidrule(r){5-6}
                           &                          & \bf{\#CLS}       & \bf{\#SUM}       & \bf{\#CLS}    & \bf{\#SUM}         \\ \midrule
\multirow{3}{*}{FR}    & Ori                 & 64          & 44,800      & 16       & 11,200        \\ \cmidrule{2-6} 
                           & Noi                    & 64          & 40,320      & 64+16    & 4,480+11,200  \\ \cmidrule{2-6} 
                           & Imb               & 64          & 40,320      & 64+16    & 4,480+4,560   \\ \midrule
\multirow{2}{*}{NYF}    & Ori                 & 212         & 33,990      & 50       & 7,010         \\ \cmidrule{2-6} 
                           & Noi                    & 212         & 30,591      & 212+50   & 3,399+7,010   \\ \midrule
\multirow{3}{*}{FR2.0} & Ori                 & 64          & 44,800      & 10       & 1,000         \\ \cmidrule{2-6} 
                           & Noi                    & 64          & 40,320      & 64+10    & 480+1,000   \\ \cmidrule{2-6} 
                           & Imb               & 64          & 40,320      & 64+10    & 480+720     \\ \bottomrule
\end{tabular}}
\label{table-statics}
\end{table}

\begin{table}[t]
\caption{The discarding probabilities for different relations.}
\centering
\scalebox{1}{
\begin{tabular}{lll}
\toprule
\textbf{Dataset}                    & \textbf{Relation ID} & \textbf{P}    \\ \midrule
\multirow{3}{*}{FewRel}    & 66-73            & 0.4  \\ \cmidrule{2-3} 
                           & 74-77            & 0.7  \\ \cmidrule{2-3} 
                           & 78-81            & 0.85 \\ \midrule
\multirow{6}{*}{FewRel2.0} & 66-68            & 0.15 \\ \cmidrule{2-3} 
                           & 69               & 0.2  \\ \cmidrule{2-3} 
                           & 70               & 0.3  \\ \cmidrule{2-3} 
                           & 71-72            & 0.35 \\ \cmidrule{2-3} 
                           & 73               & 0.4  \\ \cmidrule{2-3} 
                           & 74-75            & 0.45 \\ \bottomrule
\end{tabular}}
\label{table:dataset}
\end{table}


\subsection{Evaluation Settings }
Following previous works, we apply  instance-level evaluation metrics to evaluate the model, covering $\mathrm{B}^{3}$~\cite{bagga1998algorithms}, V-measure~\cite{rosenberg2007v} and Adjusted Rand
Index(ARI)~\cite{hubert1985comparing}. 

For quantitative validation, we divide $\mathcal{X}_N$ into $\mathcal{X}_N^{train}$ and $\mathcal{X}_N^{test}$, which account for 40\% and 60\% respectively. The active learning module selects the instance with the most information in $\mathcal{X}_N^{train}$ and trains the relation classifier. In the test phase, we merge $\mathcal{X}_K$ and $\mathcal{X}_N^{test}$, report metric scores on it. As the baselines are semi-supervised, $\mathcal{X}_N^{train}$ is also applied to the training of the baseline models to ensure a fair comparison.

For FewRel and NYT+FB, the seminal set size for Active Learning module is 32. The sample size $k$ is 32 and we sample a total of 8 epochs. In other words, a whole of 288 samples is manually labeled. As for FewRel2.0, we choose a smaller sample size: $k=8$ and keep seminal set size as 32. Finally, 96 informative samples are annotated.

\subsection{Implementation Details and Hyper-parameter Choices}
To improve the experimental effect, we use $\rm BERT_{LARGE}$ with 300M parameters in the relation representation module. We pre-train the BERT model on 3 epochs, and each epoch costs about 1 GTX 3090 GPU hour. For the discriminator, we constructed a 3-layer fully connected neural network. For active learning, $\lambda$ and $\lambda'$ are both 1. 
For optimization, different models use different optimizers. Specifically, BERT use AdamW~\cite{loshchilov2018fixing} with a learning rate of 0.00002, for discriminator, we use Adam with a learning rate of 0.0005, and for task learner of active learning, SGD is utilized. 
For baseline models, we follow their original setting without modifying any parameters except the division of the dataset.

\subsection{Main Experiment}
Table~\ref{table-main-ori} shows the quantitative evaluation results on three datasets and their variants, from which we observe that: (1) Our ARD model outperforms state-of-the-art models by a large margin. Specifically, $\mathrm{B}^3$, V-measure and ARI increased by 10.5, 13.7, and 11.4 respectively compared to OHRE on FewRel. Compared with other semi-supervised methods, the gap is even larger, rises of over 20 are achieved by ARD. This proves that ARD can efficiently discover and learn representations of novel relations at a fraction of the labor cost. (2) A universal and consistent decline in performance of baseline models from the original datasets to noisy variants and then to unbalanced variants. This demonstrates that the \textit{General OpenRE} setting is more challenging and more practical for the real scenario. The $F1$ score for RSN-CV drops dramatically from the original data to the noisy variant by 16.6. In contrast, the ARD model performs better on both the noisy and imbalanced variants than on the original dataset, even with a $F1$ score boosting by 7.2 on FewRel. This indicates the relation discovery procedure and relational active learning is robust in different scenarios. (3) The state-of-the-art models perform poorly on FewRel2.0. This is entirely to be expected, as the instances in the test set are from non-generic and low-resource domains such as biomedicine. ARD, on the other hand, still shows strong stability, confirming the cross-domain capability of the model. Further, to substantiate the applicability of our framework, we deploy ARD to a real medical dataset, as detailed in Section~\ref{appendix_section_real}.

\begin{table}[htb]
\caption{F1-measure for various active learning methods on noisy datasets.}
\centering
\scalebox{0.83}{
\begin{tabular}{clcccccc}
\toprule
\multicolumn{1}{l}{\multirow{2}{*}{\bf{Dataset}}} & \multirow{2}{*}{\bf{Model}} & \multicolumn{6}{c}{\bf{Epoch}}               \\ \cmidrule{3-8} 
\multicolumn{1}{l}{}                         &                        & \#1  & \#2 & \#3 & \#4 & \#5 & \#6 \\ \midrule
\multirow{4}{*}{FR (Noi)}    & DBAL    & 58.8 & 64.8 & 71.6 & 70.4 & 74.1 & 76.9 \\ \cmidrule{2-8} 
                           & CoreSet & 60.1 & 61.8 & 66.1 & 68.4 & 70.9 & 75.4 \\ \cmidrule{2-8} 
                           & SRAAL   & 61.9 & 64.7 & 65.7 & 69.8 & 73.7 & 73.9 \\ \cmidrule{2-8} 
                           & Ours     & \textbf{66.0} & \textbf{69.0} & \textbf{70.5} & \textbf{72.7} & \textbf{75.5} & \textbf{78.5} \\ \midrule
\multirow{4}{*}{NYF (Noi)}    & DBAL    & 47.4 & 48.6 & 51.4 & 53.3 & 54.9 & 55.5 \\ \cmidrule{2-8} 
                           & CoreSet & 45.4 & 49.5 & 52.0 & 55.3 & 56.8 & 59.2 \\ \cmidrule{2-8} 
                           & SRAAL   & 50.2 & 51.9 & 54.0 & 55.6 & 56.2 & 56.9 \\ \cmidrule{2-8} 
                           & Ours    & \textbf{56.8} & \textbf{62.5} & \textbf{66.6} & \textbf{68.3} & \textbf{69.3} & \textbf{69.9} \\ \midrule
\multirow{4}{*}{FR2.0 (Noi)} & DBAL    & 46.9 & 50.5 & 51.4 & 51.7 & 52.2 & 53.7 \\ \cmidrule{2-8} 
                           & CoreSet & 44.0 & 45.5 & 50.3 & 51.9 & 53.0 & 54.2 \\ \cmidrule{2-8} 
                           & SRAAL   & 45.0 & 49.7 & 51.8 & 52.0 & 52.8 & 53.9 \\ \cmidrule{2-8} 
                           & Ours    & \textbf{48.8} & \textbf{51.2} & \textbf{52.4} & \textbf{53.2} & \textbf{53.5} & \textbf{54.5} \\ \bottomrule
\end{tabular}}
\label{table-active-baseline}
\end{table}

\begin{table*}[h]
\centering
\caption{ Main results on three original datasets and their variants. Ori, Noi, and Imb stand for original, noisy and imbalanced respectively. Ori corresponds to the conventional setting. Noi, and Imb refer to the general setting. Results are the average of 3 experiments with different random seeds.}
\scalebox{1}{
\begin{tabular}{l|l|ccc|ccc|c}
\toprule[1.5pt]
\multirow{2}{*}{\bf{Dataset}}   & \multicolumn{1}{c|}{\multirow{2}{*}{\bf{Model}}} & \multicolumn{3}{c|}{$\mathrm{\bf{B}}^{\bf{3}}$} & \multicolumn{3}{c|}{\bf{V-measure}} & \multirow{2}{*}{\bf{ARI}} \\ \cmidrule{3-8}
                           & \multicolumn{1}{c|}{}                       & F1    & Prec.  & Rec.  & $V$       & Hom.    & Comp.    &                      \\ \midrule
\multirow{4}{*}{FR (Ori)}    & RSN-CV                 & $59.2_{\pm0.5}$  & 55.5   & 63.4  & $72.9_{\pm0.3}$     & 69.1     & 77.2    &  $47.2_{\pm0.9}$                 \\ \cmidrule{2-9} 
                           & SelfORE                                    & $60.6_{\pm0.6}$  & 60.2   & 61.1  & $68.4_{\pm1.2}$     & 67.5     & 69.3    & $56.0_{\pm0.7}$                 \\ \cmidrule{2-9} 
                           & OHRE                                       & $63.1_{\pm1.0}$  & 54.9   & 74.1  & $71.4_{\pm0.8}$     & 64.9     & 79.4    & $52.7_{\pm1.0}$                 \\ \cmidrule{2-9} 
                           & Ours                                       & \bm{$73.6_{\pm0.8}$}  & \textbf{70.7}   & \textbf{76.8}  & \bm{$85.1_{\pm0.1}$}     & \textbf{84.9}     & \textbf{85.3}    & \bm{$64.1_{\pm0.8}$}                 \\ \midrule
\multirow{4}{*}{NYF (Ori)}    
& RSN-CV                 & $49.7_{\pm0.8}$  & 39.3   & 67.6  & $64.2_{\pm0.8}$     & 56.7     & \textbf{73.9}    & $35.1_{\pm0.2}$                 \\ \cmidrule{2-9} 
& SelfORE           & \bm{$54.9_{\pm0.7}$}  & \textbf{52.8}   & 57.2  & \bm{$72.5_{\pm0.3}$}     & 71.6     & 73.4    & \bm{$56.6_{\pm0.6}$}                 \\ \cmidrule{2-9} 
& OHRE          & $41.2_{\pm0.2}$  & 28.5   & \textbf{74.3}  & $54.1_{\pm0.5}$     & 42.7     & 73.7    & $26.5_{\pm0.6}$                 \\ \cmidrule{2-9} 
& Ours         & $51.4_{\pm0.5}$  & 45.0   & 60.0  & $72.3_{\pm0.4}$     & \textbf{75.0}     & 69.8    & $45.1_{\pm0.7}$                 \\ \midrule
\multirow{4}{*}{FR2.0 (Ori)} & RSN-CV                 & $27.7_{\pm1.2}$  & 18.2   & 58.2  & $48.8_{\pm0.2}$     & 39.5     & 63.7    & $13.4_{\pm0.7}$                 \\ \cmidrule{2-9} 
                           & SelfORE                                    & $36.7_{\pm0.8}$  & 26.2   & \textbf{61.3}  & $60.2_{\pm0.8}$     & 52.2     & \textbf{71.2}    & $27.3_{\pm0.3}$                 \\ \cmidrule{2-9} 
                           & OHRE                                       & $25.1_{\pm0.6}$  & 18.5   & 38.9  & $15.8_{\pm0.3}$     & 14.4     & 17.6    & $9.4_{\pm0.7}$                 \\ \cmidrule{2-9} 
                           & Ours                                       & \bm{$48.8_{\pm0.7}$}  & \textbf{43.2}   & 56.1  & \bm{$65.1_{\pm0.8}$}     & \textbf{60.8}     & 70.1    & \bm{$34.4_{\pm0.8}$}                 \\ \midrule[1.5pt]
\multirow{4}{*}{FR (Noi)}    & RSN-CV                 & $42.6_{\pm0.7}$  & 30.1   & 72.6  & $66.6_{\pm0.3}$     & 56.4     & 81.2    & $28.3_{\pm0.3}$                 \\ \cmidrule{2-9} 
                           & SelfORE                                    & $51.3_{\pm0.8}$  & 49.3   & 53.5  & $56.4_{\pm0.5}$     & 55.2     & 57.7    & $45.8_{\pm0.4}$                 \\ \cmidrule{2-9} 
                           & OHRE                                       & $32.5_{\pm0.2}$  & 20.4   & 79.8  & $57.5_{\pm1.2}$     & 46.1     & 76.3    & $26.3_{\pm1.2}$                 \\ \cmidrule{2-9} 
                           & Ours                                       & \bm{$80.8_{\pm0.9}$}  & \textbf{75.7}   & \textbf{86.7}  & \bm{$90.2_{\pm0.5}$}     & \textbf{89.0}     & \textbf{91.4}    & \bm{$71.3_{\pm0.5}$}                 \\ \midrule
\multirow{4}{*}{NYF (Noi)}    & RSN-CV                 & $43.0_{\pm0.5}$  & 30.6   & 72.3  & $59.6_{\pm0.1}$     & 50.7     & 72.3    & $30.6_{\pm0.7}$                 \\ \cmidrule{2-9} 
                           & SelfORE                                    & $48.6_{\pm1.2}$  & 43.8   & 54.6  & $65.7_{\pm0.1}$     & 65.1     & 66.3    & $46.2_{\pm0.3}$                 \\ \cmidrule{2-9} 
                           & OHRE                                       & $36.4_{\pm0.7}$  & 25.1   & 66.2  & $48.1_{\pm0.3}$     & 38.2     & 64.6    & $30.2_{\pm0.9}$                 \\ \cmidrule{2-9} 
                           & Ours                                       & \bm{$71.3_{\pm0.2}$}  & \textbf{60.8}   & \textbf{86.2}  & \bm{$72.9_{\pm0.3}$}     & \textbf{70.5}     & \textbf{75.5}    & \bm{$51.0_{\pm0.3}$}                 \\ \midrule
\multirow{4}{*}{FR2.0 (Noi)} & RSN-CV                 & $27.7_{\pm0.2}$  & 32.4   & 24.2  & $31.4_{\pm0.4}$     & 30.1     & 32.8    & $10.2_{\pm0.5}$                 \\ \cmidrule{2-9} 
                           & SelfORE                                    & $32.8_{\pm1.1}$  & 24.7   & 48.9  & $54.8_{\pm1.0}$     & 47.3     & 65.1    & $27.1_{\pm1.0}$                 \\ \cmidrule{2-9} 
                           & OHRE                                       & $25.2_{\pm0.5}$  & 15.9   & 60.5  & $50.0_{\pm1.2}$     & 40.1     & 66.4    & $16.2_{\pm0.2}$                 \\ \cmidrule{2-9} 
                           & Ours                                       & \bm{$55.0_{\pm0.7}$}  & \textbf{52.8}   & \textbf{57.4}  & \bm{$69.3_{\pm0.7}$}     & \textbf{65.1}     & \textbf{74.0}    & \bm{$38.5_{\pm0.9}$}                 \\ \midrule[1.5pt]
\multirow{4}{*}{FR (Imb)}    & RSN-CV                 & $37.2_{\pm1.0}$  & 24.6   & 76.2  & $65.5_{\pm0.6}$     & 55.1     & 80.8    & $25.3_{\pm0.5}$                 \\ \cmidrule{2-9} 
                           & SelfORE                                    & $48.3_{\pm0.2}$  & 44.2   & 53.5  & $53.7_{\pm0.1}$     & 56.4     & 51.3    & $44.2_{\pm1.3}$                 \\ \cmidrule{2-9} 
                           & OHRE                                       & $31.0_{\pm1.0}$  & 19.8   & 71.1  & $56.1_{\pm0.6}$     & 44.1     & 77.6    & $22.6_{\pm1.0}$                 \\ \cmidrule{2-9} 
                           & Ours                                       & \bm{$76.5_{\pm0.6}$}  & \textbf{74.7}   & \textbf{78.4}  & \bm{$86.5_{\pm0.7}$}     & \textbf{86.8}     & \textbf{86.2}    & \bm{$67.8_{\pm0.6}$}                 \\ \midrule
\multirow{4}{*}{FR2.0 (Imb)}    & RSN-CV              & $26.4_{\pm0.4}$  & 20.6   & 36.8  & $31.5_{\pm0.3}$     & 25.2     & 41.9    & $22.7_{\pm0.7}$                 \\ \cmidrule{2-9} 
                           & SelfORE                                    & $31.3_{\pm1.2}$  & 22.4   & 52.2  & $52.9_{\pm0.4}$     & 45.8     & 62.6    & $25.7_{\pm0.9}$                 \\ \cmidrule{2-9} 
                           & OHRE                                       & $22.6_{\pm0.6}$  & 13.9   & 60.7  & $45.5_{\pm1.0}$     & 35.4     & 63.6    & $13.4_{\pm0.6}$                 \\ \cmidrule{2-9} 
                           & Ours                                       & \bm{$52.4_{\pm0.6}$}  & \textbf{50.1}   & \textbf{54.9}  & \bm{$67.4_{\pm0.2}$}     & \textbf{63.2}     & \textbf{72.2}    & \bm{$36.4_{\pm0.3}$}                 \\ \bottomrule[1.5pt]
\end{tabular}}
\label{table-main-ori}
\end{table*}

\begin{figure}[htb]
    \centering
    \scalebox{1}{\includegraphics[width = 0.98 \linewidth]{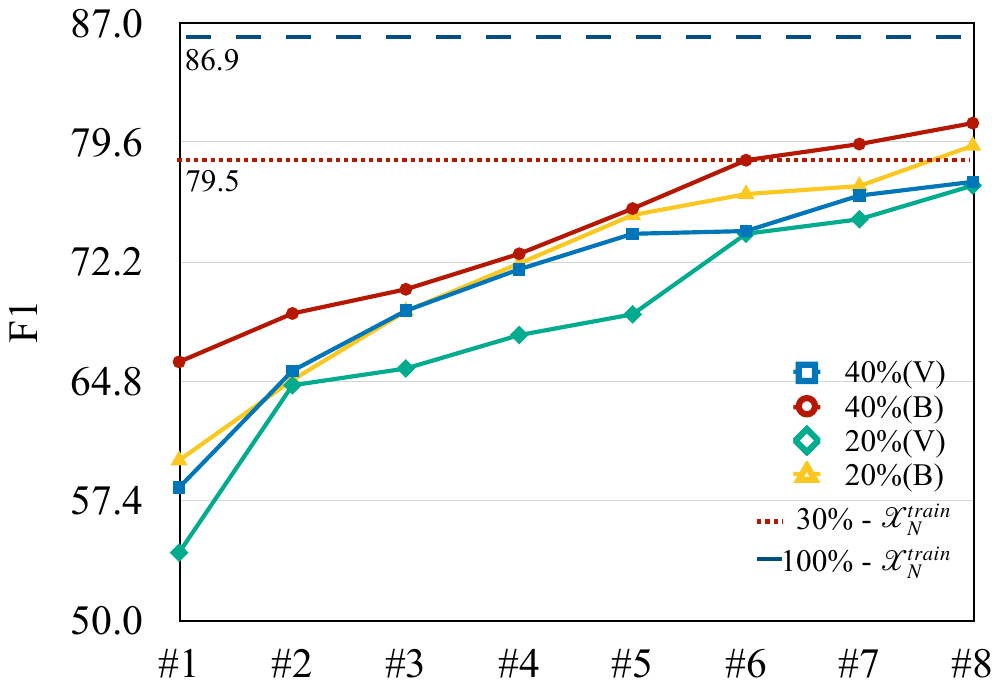}}
    \caption{F1-measure on noisy FewRel, (V) denotes the $\beta$-VAE and (B) denotes the BERT encoder.}
    \label{fig:ablation}
\end{figure}

\subsection{Analysis on Active Learning}
\textbf{The Efficiency of Active Learing }
Table~\ref{table-active-baseline} shows the results of our active learning approach compared to various active learning baseline models including DBAL~\cite{gal2017deep}, CoreSet~\cite{sener2018active}, SRAAL~\cite{zhang2020state}. It can be observed that in each iteration, our model outperforms the other models, indicating that our method can consistently sample informative samples.
In particular, our method performs significantly better on the NYT-FB dataset where the category count is much larger. Compared to the baseline models, our method ensures the information content and category diversity of the selected samples by enabling samples of the present batch to interact implicitly with previously selected samples through the discriminator.
Besides, we report first 8 cases selected by discriminator which are regarded as the most representative instances(instances with novel relations) in Section~\ref{appendix_section_case}.

\noindent\textbf{The Impact of Different Encoder and Scope of Query }
Figure~\ref{fig:ablation} shows the experimental results on noisy FewRel with different encoders and query ranges. The ``query ranges'' represents the ratio of $\mathcal{X}_{N}^{train}$ to $\mathcal{X}_N$, 
We also explore the impact of $\beta$-VAE~\cite{higgins2016beta} and BERT as encoders. 
From the results we observe that: (1) 
Generally, the model performance is proportional to the size of $\mathcal{X}_{N}^{train}$. However, the results are improved marginally as the number of samples increase. 
But the model still yields better performance when the query range is 40\%. (2) The comparisons between the VAE and the BERT encoder are in line with intuition. Although VAE is intuitive and can be more easily trained, BERT still shows superiority in empirical results. 

\noindent\textbf{Compare with Manual Random Selection.}
In Figure~\ref{fig:ablation}, the gain from 288 informative instances (approximately 8\% of $\mathcal{X}_{N}^{train}$) selected by the active learning is similar to the gain from 30\% of instances randomly selected. When trained with the full amount of $\mathcal{X}_{N}^{train}$, the $F1$ is 6.1\% higher than ARD while costing 12 times as much in human effort.

\noindent\textbf{The Impact of Different Sampling Strategies }
In order to prove the effectiveness of the active learning method, we conduct a further ablation experiment. As mentioned above, our sampling strategy is to select the $k$ instances with the highest confidence for manual labeling. In the ablation experiment, we test two other sampling strategies: selecting the $k$ instances with the lowest confidence; randomly selecting $k$ instances by human. The comparison results are shown in Table~\ref{table:sample}.

It can be seen that after being trained by instances with the highest confidence, the model achieves the most improvement. 
In contrast, instances with the lowest confidence contribute very little to improving the performance of the model. 
Even with the continuous increase of training data, the improvement is extremely little. 
The results prove that the active learning model does select the most informative instances. 

\begin{table}[ht]
\caption{Comparisons of F1-measure between different sampling strategies on noisy FewRel dataset.}
\centering
\scalebox{1}{
\begin{tabular}{cccc}
\toprule
\bf{Epoch}            & \bf{Lowest}        & \bf{Random}     & \bf{Highest}  \\ \midrule
\#1                    & 57.1      & 58.7      & 66.0      \\ 
\#2                    & 57.1      & 60.4      &  69.0      \\ 
\#3                    & 57.8      & 65.6      &  70.5    \\ 
\#4                    & 57.6      & 67.2      &  72.7      \\ 
\#5                    & 57.6      & 67.7      &  75.5      \\ 
\#6                    & 58.1      & 67.7      &  78.5      \\ \bottomrule
\end{tabular}}
\label{table:sample}
\end{table}

\noindent\textbf{Time Efficiency of Relational Active Learning}
In practice, it is often the time spent on manual annotation that is the time-consuming bottleneck. Nevertheless, the sampling strategy for active learner should also select samples in a time-efficient manner as much as possible. We analysis the time efficiency of different active learning methods. Table~\ref{table:active_time} shows the average time for different methods to sample once on the corresponding dataset. DBAL is the most competitive baselines in terms of their achieved mean time efficiency. Our method fell marginally behind DBAL, however, our method is outperformed in accuracy by all other methods. 

\begin{table}[h]
\centering
\caption{Average time token to sample once on the corresponding dataset.}
\scalebox{1}{
\begin{tabular}{lllll}
\toprule
\multirow{2}{*}{\textbf{Dataset}} & \multicolumn{4}{c}{\textbf{Time(ms)}}    \\ \cmidrule{2-5} 
                         & DBAL  & CoreSet & SRAAL & Ours  \\ \midrule
FewRel                   & 157.0 & 1145.2   & 409.3 & 465.1 \\ \midrule
NYT+FB                   & 157.9 & 74.3   & 132.7 & 101.9 \\ \midrule
FewRel2.0                & 181.4 & 209.4   & 418.4 & 9.2 \\ \midrule
Average                  & \textbf{165.4} & 476.3   & 320.1 & \underline{192.0} \\ \bottomrule
\end{tabular}}
\label{table:active_time}
\end{table}

\subsection{Analysis on Relational Outlier Detection}

\noindent\textbf{The Effects of Relational Outlier Detection}
ARD employs novel relation discovery module to distinguish between known and novel relations, preserving the active learning module to more efficiently select informative novel relations without being distracted by known relations. To demonstrate the effectiveness and significance of the novel relation discovery, we perform ablation experiments over LOF algorithm on three noisy variants. Table~\ref{table-outlier} shows the experimental results, and we note that: (1) Despite the robust learning ability of active learning on novel relations, the model performances show different degrees of degradation after the removal of the LOF algorithm. (2) Average of decline of $F1$ scores in each epoch on the FewRel, NYT+FB, and FewRel2.0 datasets is 2.82, 8.02, 6.36 respectively, with the most severe drop on NYT+FB. The phenomenon is intuitive, as the NYT+FB dataset contains the most known relations; the more noise (known relations) there is, the more confused the active learning module becomes about the novel relations. The results demonstrate the novel relation discovery module plays a key role as ``noise reduction''. 

\noindent\textbf{The Impact of Different Outlier Detection Algorithms} 
We compare LOF with two different algorithms for the relational outlier detection, including IsolationForest~\cite{liu2008isolation}, and OneClassSVM~\cite{scholkopf2001estimating}. 
We evaluate the F1-measure of these three algorithms solely on the discovery of novel relations, the results are reported in Table~\ref{table:lof}. Our LOF algorithm outperforms by large margins, achieving 83.9\% F1-measure on FewRel dataset.
The principle of the IsolationForest algorithm is to cut data points and isolate data points one by one. Thus the data needs more cuts to be isolated. 
The main reason for the poor performance of this algorithm is a large amount of the test data. For the same type of new relations, their distribution is relatively dense, and the number of cuts will also increase. Moreover, the dimensions of relation representation are 2048, while IsolationForest has poor processing capabilities for high-dimensional features. Hence, it yields relevant poor results. OneClassSVM aims to learn a tight decision boundary from normal data and treats points outside the decision boundary as abnormal points. In the relational feature space, the distribution of known relations and novel relations are complicated.
Thus the OneClassSVM is likely to learn an over-fitting decision boundary, resulting in poor performance.

\begin{table}[ht]
\caption{Ablation experiments over novel relation discovery module on noisy datasets.}
\centering
\scalebox{0.88}{
\begin{tabular}{llccccc}
\toprule
\multicolumn{1}{l}{\multirow{2}{*}{\bf{Dataset}}} & \multirow{2}{*}{\bf{Model}} & \multicolumn{5}{c}{\bf{Epoch}}               \\ \cmidrule{3-7} 
\multicolumn{1}{l}{}        &   & \#1  & \#2 & \#3 & \#4 & \#5   \\ \midrule
\multirow{2}{*}{FR (Noi)}    & ARD & 66.0 & 69.0 & 70.5 & 72.7 & 75.5 \\ \cmidrule{2-7} 
                           & w/o LOF  & 62.8 & 64.4 & 68.5 & 70.5 & 73.4 \\ \midrule
\multirow{2}{*}{NYF (Noi)}    & ARD & 56.8 & 62.5 & 66.6 & 68.3 & 69.3 \\ \cmidrule{2-7} 
                           & w/o LOF    & 47.7 & 53.8 & 57.2 & 60.3 & 64.4 \\ \midrule
\multirow{2}{*}{FR2.0 (Noi)} & ARD & 48.8 & 51.2 & 52.4 & 53.2 & 53.5 \\ \cmidrule{2-7} 
                           & w/o LOF    & 42.9 & 44.2 & 45.6 & 46.4 & 48.2 \\ \bottomrule
\end{tabular}}
\label{table-outlier}
\end{table}

\begin{table}[ht]
\caption{F1-measure on noisy FewRel and FewRel2.0 with different outlier detection algorithms.}
\centering
\scalebox{1}{
\begin{tabular}{lccc}
\toprule
\multirow{2}{*}{\textbf{Dataset}} & \multicolumn{3}{c}{\textbf{Method}} \\ \cmidrule{2-4} 
                         & IF    & OneClassSVM & LOF  \\ \midrule
FewRel                   & 64.0  & 47.3        & \textbf{83.9} \\ \midrule
FewRel2.0                & 63.1  & 54.1        & \textbf{80.3} \\ \bottomrule
\end{tabular}}
\label{table:lof}
\end{table}

\subsection{Case Study of Active Learning}
\label{appendix_section_case}
As shown in Table~\ref{table:case}, we report 8 cases selected by discriminator in the first iteration on noisy FewRel2.0 dataset, where 64 relations are pre-trained and seen. With the highest confidence, the discriminator successfully selects sentences with unseen relations and guarantees the diversity of relation categories.

\subsection{Additional Exploration}
\subsubsection{Visualization of Relation Representations}
In order to intuitively demonstrate the distribution of novel relations relative to known relations and, on the other hand, to illustrate the benefits of introducing contrastive loss, we visualize the relation representation $\boldsymbol{h}_{r}(x)$ after dimension reduction using $t$-SNE~\cite{maaten2008visualizing}. 

\begin{figure}[h]
\centering
\subfigure[Train using only traditional cross-entropy loss.] { \label{tsne1} 
\includegraphics[height = 0.45 \columnwidth,width=0.46\columnwidth]{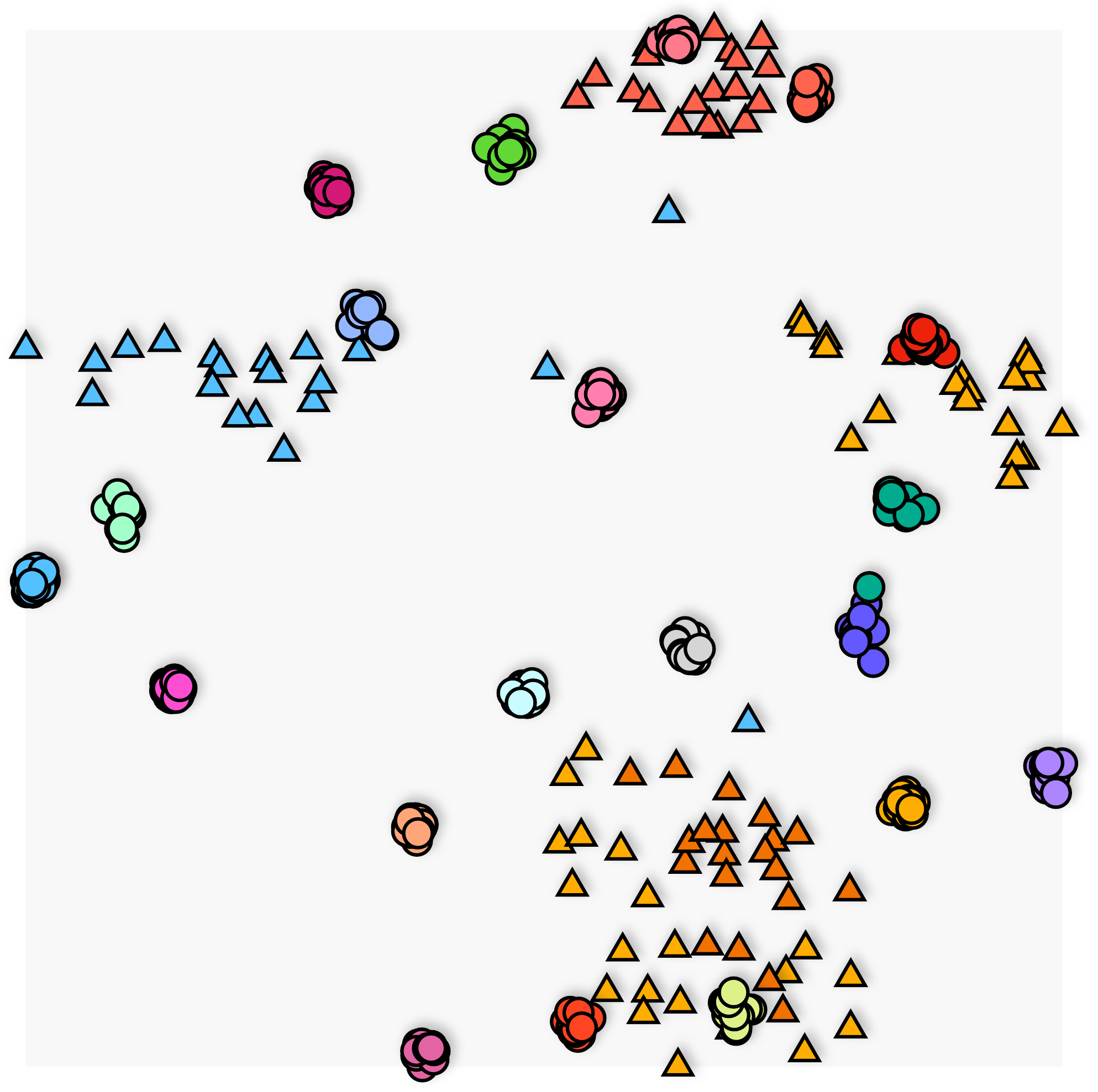} 
} 
\subfigure[Plus contrastive loss.] 
{ \label{tsne2} 
\includegraphics[height = 0.45 \columnwidth, width=0.46\columnwidth]{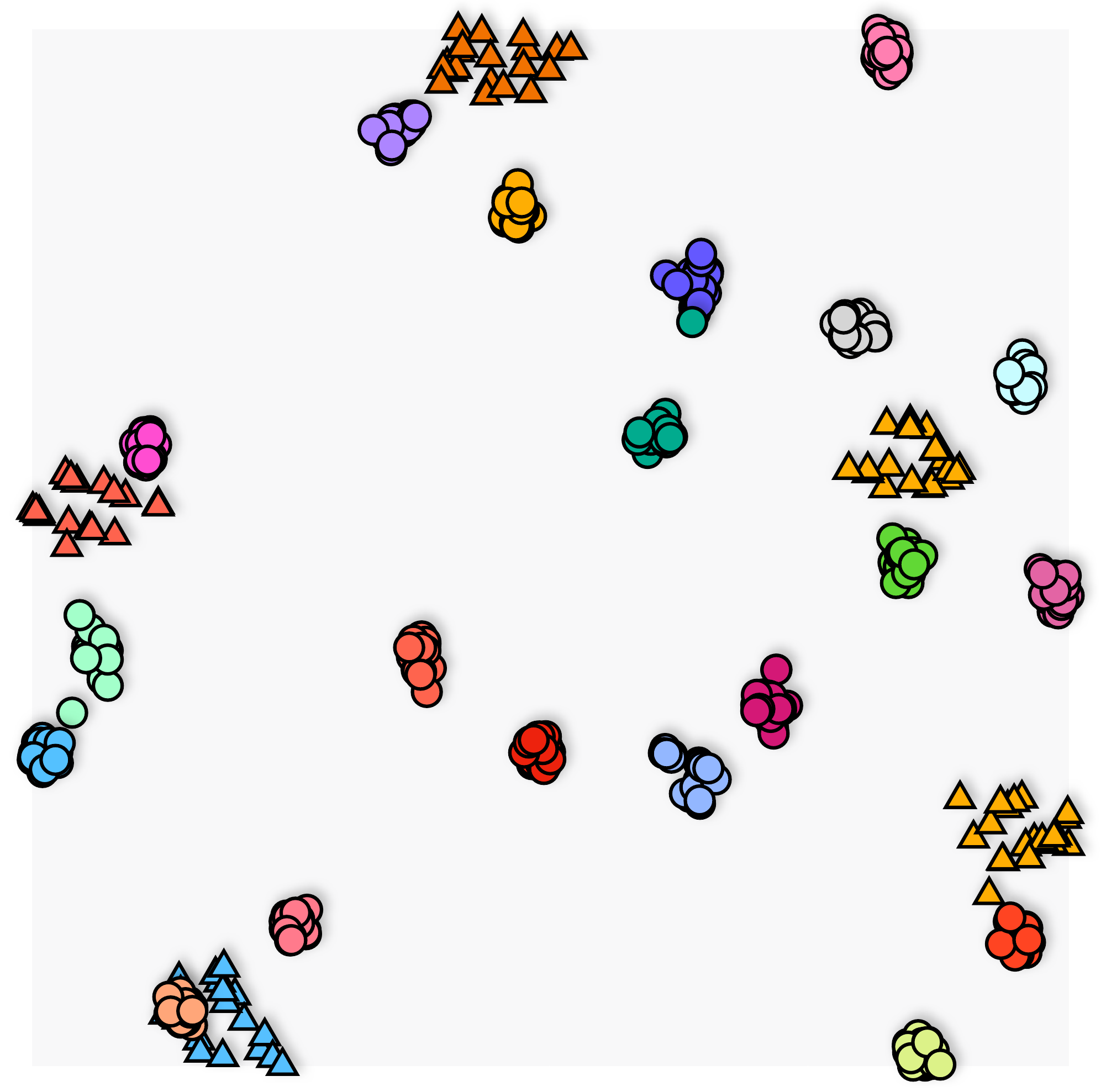} 
} 
\caption{$t$-SNE visualization of relation representation. The known and novel relations are distinguished by circular and triangular symbols respectively.} 
\label{tsne} 
\end{figure}

As illustrated in Figure~\ref{tsne}, instances of the same known relation type are densely clustered with a high local density, while instances of novel relations distribute dispersedly. This fact strongly supports the premise of the LOF algorithm. Also, comparing subfigures~\ref{tsne1} and~\ref{tsne2}, we observe that contrastive loss firmly constrains the distribution of intra-class instances. In pre-experiments on FewRel, the introduction of contrastive loss boosts the accuracy in distinguishing known and novel relations from 79.3\% to 83.9\%.


\subsubsection{Performance of our Relation Representation on Supervised RE}
\label{appendix_section_supervisre}
To demonstrate the effectiveness of the relation representation described in the Methodology section, we conduct a series of experiments on supervised RE task.
First, we conduct extensive experiments on the biomedical relation extraction benchmark DDI'13~\cite{herrero2013ddi} and show in Table \ref{tab:ddi}. We make comparisons with various previous state-of-the-art methods, which fall into two groups according to the neural network architecture: convolutional neural network (CNN) based methods and recurrent neural network (RNN) based methods. For the first group, we report the results of SCNN~\cite{zhao2016drug}, CNN-bioWE~\cite{liu2016drug} and MCCNN~\cite{quan2016multichannel}, which uses syntax word embeddings, biomedical-related embeddings and multi-channel word embeddings for feature extraction, respectively. For recurrent based networks, we report the reults of Joint AB-LSTM~\cite{sahu2018drug}, Position-aware LSTM~\cite{zhou2018position}, RvNN~\cite{lim2018drug} and BERE~\cite{hong2020novel}. Joint AB-LSTM jointly trains two bidrectional LSTM (Bi-LSTM) with different pooling mechanisms: max-pooling for one Bi-LSTM and attentive pooling for the other. Position-aware LSTM adopt position information as attention mechanism for the training of LSTM. RvNN and BERE incorporates parse-tree information to enhance the performance of prediction. Each model is trained on the training dataset to predict a relation class of five pre-defined relation types for the input sequence.

\begin{table}[t]
\caption{\label{tab:ddi} Results on DDI'13 dataset. The first seven rows are the results of the previous SOTA methods, and the bottom results are for ours method in supervised relation learning.}
\centering
\scalebox{1}{
\begin{tabular}{lccc}
\toprule
\textbf{Methods} & \textbf{Pre.} & \textbf{Rec.} & \textbf{F1}  \\
\midrule
SCNN & 69.1 & 65.1 & 67.0\\
CNN-bioWE & 75.7 &64.7 &69.8 \\
MCCNN & 75.9 & 65.2 &70.2 \\
Joint AB-LSTM & 73.4 & 69.6 & 71.5\\
RvNN & 74.4 & 69.3 & 71.7 \\
Position-aware LSTM & 75.8 & 70.4 & 73.0 \\
BERE & 76.8 & 71.3 & 73.9 \\ \midrule
Ours & \textbf{92.3} & \textbf{84.4}  & \textbf{86.8} \\ \bottomrule
\end{tabular}}
\end{table}

\begin{figure}[t]
\centering
\scalebox{1}{
\includegraphics[width=0.9\linewidth]{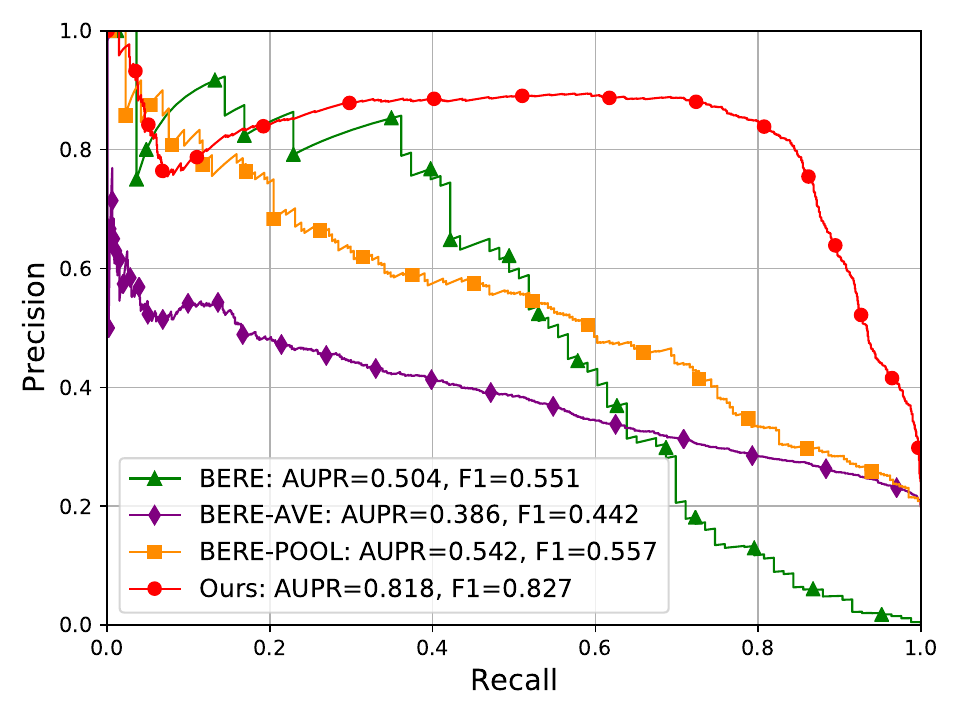}}
\caption{Precision-recall curve of BERE and our model. }
\label{fig:pr}
\end{figure}

To further evaluate the performance of our representation method on large-scale distantly annotated dataset, we conduct another set of experiments on the DTI dataset. As on the DTI dataset,  previous literature has shown the superiority of BERE compared with CNN-based and RNN-based baselines, we mainly take BERE as the baseline of our experiments. For fairness, we follow the settings of BERE by using precision-recall curve, the area under the precision-recall curve and the $F_1$ score as the evaluation metrics. We re-run the open-source code of BERE and its two variants: BERE-AVE, BERE-POOL. BERE-AVE adopt the average pooling mechanism to aggregate the semantic information over instances in a bag. BERE-POOL uses the max-pooling strategy. The implementation details of our model on the DTI dataset are identical to the DDI'13 dataset. The precision-recall curve is shown in Figure~\ref{fig:pr}, which indicates the significant performance of our representation method. 

\begin{table*}[ht]
\caption{Cases selected by the confidence score of the discriminator and the novel relations, where \textit{\color{red}{red}} and \textit{\color{blue}{blue}} represent the head and tail entities}. 
\centering
\scalebox{1}{
\begin{tabular}{p{11.5cm}l}
\toprule
\textbf{Selected sentence} & \textbf{Novel relation} \\ \midrule

Ectopic overexpression of mir-497 promotes chemotherapy resistance in glioma cells by targeting \textit{\color{red}{pdcd4}}, a tumor suppressor that is involved in \textit{\color{blue}{apoptosis}}. &\textit{Biological process involves gene product} \\ \midrule 

As full-length bid is a weaker apoptogen than \textit{\color{red}{tbid}},we propose that the phosphorylation of bid by jnks, followed by the accumulation of the full-length protein, delays attainment of \textit{\color{blue}{apoptosis}}, and allows the cell to evaluate the stress and make a decision regarding the response strategy. & \textit{Biological process involves gene product} \\ \midrule

Pretreatment with dexamethasone 1 hour before \textit{\color{red}{cyclophosphamide injection}} significantly down-regulated \textit{\color{blue}{cyclophosphamide}} induced bladder nuclear factor-u03bab dependent luminescence, ameliorated the grossly evident pathological features of acute inflammation and decreased cellular immunostaining for nuclear factor-u03bab in the bladder. & \textit{Ingredient of} \\ \midrule

Trastuzumab emtansine (\textit{\color{red}{t-dm1}}), an antibody-drug conjugate comprising the cytotoxic agent dm1, a stable linker, and \textit{\color{blue}{trastuzumab}}, has demonstrated substantial activity in human epidermal growth factor receptor 2 (her2), -positive metastatic breast cancer, raising interest in evaluating the feasibility and cardiac safety of t-dm1 in early-stage breast cancer (ebc). & \textit{Ingredient of} \\ \midrule

Here we looked for evidence of adult hippocampal \textit{\color{red}{neurogenesis}} using immunohistochemical techniques for the endogenous marker doublecortin (\textit{\color{blue}{dcx}}) in 10 species of microchiropterans euthanized and perfusion fixed at specific time points following capture. & \textit{Gene plays role in process} \\ \midrule

Here, we explored the effects of the novel class ii-specific "histone deacetylase inhibitors (hdacis) mc1568 and mc1575 on interleukin-8 (il-8) expression and \textit{\color{red}{cell proliferation}} in cutaneous melanoma cell line \textit{\color{blue}{gr}} -m and uveal melanoma cell line ocm-3 upon stimulation with phorbol 12-myristate 13-acetate (pma). &   \textit{Gene plays role in process}  \\ \midrule

Data indicate that the structurally disordered and abnormally formed ecm of \textit{\color{red}{uterine fibroids}} contributes to \textit{\color{blue}{fibroid}} formation and growth. & \textit{Classified as} \\ \midrule

however, individuals heterozygous for both beta "e", "and", beta thalassaemia (hbe/\textit{\color{red}{beta thalassaemia}}) have a severe clinical disorder which in some cases may approach that seen in \textit{\color{blue}{homozygous beta thalassaemia}} and which is by far the commonest form of symptomatic thalassaemia in the indian subcontinent and south-east asia. & \textit{Classified as}
\\ \bottomrule

\end{tabular}}
\label{table:case}

\end{table*}

\subsubsection{Impact of The Size of BERT Model}
We change the size of $\rm{BERT}$ in ARD. The results of this ablation experiment are shown in the Table \ref{table-bertsize}. We can find that the size of $\rm{BERT}$ is not the key factor to bring gain, and even if we use $\rm{BERT_{BASE}}$ as the backbone, the performance of ARD is still considerably higher than that of the baseline model using $\rm{BERT_{LARGE}}$ in Table \ref{table-main-ori}.

\begin{table}[t]
\caption{Ablation experiments over the size of BERT model on FewRel datasets.}
\centering
\scalebox{1}{
\footnotesize
\renewcommand\arraystretch{1}
\setlength\tabcolsep{2.8pt}
\begin{tabular}{c|l|ccc|ccc|c}
\toprule[1.5pt]
\multirow{2}{*}{\bf{\shortstack{Data\\-set}}}   & \multicolumn{1}{c|}{\multirow{2}{*}{\bf{Model}}} & \multicolumn{3}{c|}{$\mathrm{\bf{B}}^{\bf{3}}$} & \multicolumn{3}{c|}{\bf{V-measure}} & \multirow{2}{*}{\bf{ARI}} \\ \cmidrule{3-8}
                           & \multicolumn{1}{c|}{}                       & F1    & Prec.  & Rec.  & $V$       & Hom.    & Comp.    &                      \\ \midrule
\multirow{2}{*}{\shortstack{FR\\(Ori)}}  
                           & BASE                                       & 68.8  & 55.7   & 90.0  & 79.2     & 73.3     & 86.2    & 55.5                 \\ \cmidrule{2-9} 
                           & LARGE                                       & 73.6  & 70.7   & 76.8  & 85.1     & 84.9     & 85.3    & 64.1                 \\ \midrule
\multirow{2}{*}{\shortstack{FR\\(Noi)}}    
                           & BASE                                       & 75.7  & 66.1   & 88.8  & 82.2     & 78.5     & 86.2    & 62.2                 \\ \cmidrule{2-9} 
                           & LARGE                                       & 80.8  & 75.7   & 86.7  & 90.2     & 89.0     & 91.4    & 71.3                 \\ \midrule
\multirow{2}{*}{\shortstack{FR\\(Imb)}}
                           & BASE                                       & 70.6  & 63.4   & 57.5  & 81.2     & 74.2     & 68.4    & 59.3                 \\ \cmidrule{2-9} 
                           & LARGE                                       & 76.5  & 74.7   & 78.4  & 86.5     & 86.8     & 86.2    & 67.8                 \\  \bottomrule[1.5pt]
\end{tabular}}
\label{table-bertsize}
\end{table}

\subsection{Practical Application on Real-world Dataset}
\label{appendix_section_real}
\begin{figure}[t]
\centering
\scalebox{1}{
\includegraphics[width=0.9\linewidth]{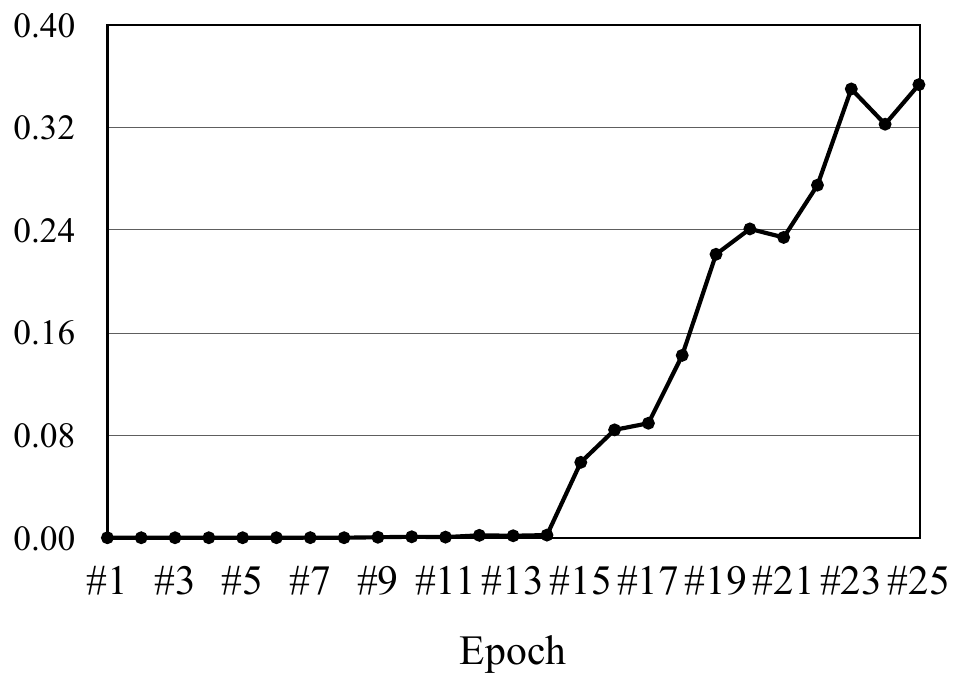}}
\caption{Discriminator's confidence for $\mathcal{X}_U$ in each epoch.}
\label{fig:confidence}
\end{figure}

We apply the ARD framework in real-world scenarios to verify its practicability. With the increasing number of publications about COVID-19, it is a challenge to extract personalized knowledge suitable for each researcher~\cite{dong2021survey, huang2022towards}. ~\cite{barros2020covid} aims to build a new semantic-based pipeline for recommending biomedical entities to scientific researchers. In this work, the researchers utilize MER~\cite{couto2018mer} as NER annotation server. As a result, 9,000 articles are automatically annotated with relevant items/concepts for COVID-19. And for further relation extraction task, due to the expensive manual annotation costs, the researchers merely take initial steps towards the results, providing a small sample dataset of ten documents, with all possible relationships between the four types of entities identified by NER pipeline. Thus, we
were able to establish ten different types of relations, encompassing the four ontologies (CHEBI, DO, HPO, and GO). We follow the relation types, and apply ARD framework in the results. We take sample size $k$ of 200 and sample 25 epochs. Finally, a total of 139,479 relations between entity pairs are automatically obtained by ARD. The statistical results of the data are shown in Table~\ref{table:application}. 
We also report the confidence of the discriminator in each epoch for $\mathcal{X}_U$. As can be observed from the Figure~\ref{fig:confidence}, the confidence is progressively increasing as the training epoch increases, which indicates that the model is becoming more confident in the classification results. In an ideal case, the confidence should converge toward 0.5.

\begin{table}[t]
\centering
\caption{Statistical results of dataset for COVID-19.}
\scalebox{1}{
\begin{tabular}{ll}
\toprule
\textbf{Relation}    & \textbf{Count} \\ \midrule
CHEBI-CHEBI & 4680  \\
CHEBI-HP    & 20455 \\
GO-HP       & 17254 \\
DO-DO       & 14430 \\
CHEBI-DO    & 2415  \\
HP-HP       & 48236 \\
HP-DO       & 19770 \\
GO-CHEBI    & 1285  \\
GO-DO       & 3615  \\
GO-GO       & 7303  \\ \bottomrule
\end{tabular}}
\label{table:application}
\end{table}

\section{Conclusion and Future Work}

The paper proposes Active Relation Discovery (ARD), which aims at accurately discovering and meaningfully annotating new semantic relations under the \textit{General OpenRE} setting. 
By introducing outlier detection and active learning, ARD solves two problems: (1) \textit{Sufficient capabilities to distinguish between known and novel relations}, with robust performance under General OpenRE settings. (2) \textit{Avoiding Secondary labeling of downstream tasks}.
Extensive experiments are conducted to demonstrate the effectiveness of ARD.

As a pioneering work in OpenRE, several directions can be further explored: (1) Better methods to discriminate and annotate novel relations in \textit{General OpenRE} setting.
(2) Better methods to capture the core relational features for relation representation.  
(3) Combination with bootstrapping methods to partially replace active learning. (4) Combination with lifelong learning to continuously incorporate novel relations. 
(5) A universal schema for the standard of active relation learning. 

\section*{Acknowledgement}
This research is supported by National Natural Science Foundation of China (Grant No.62276154 and 62011540405), Beijing Academy of Artificial Intelligence (BAAI), the Natural Science Foundation of Guangdong Province (Grant No. 2021A1515012640), Basic Research Fund of Shenzhen City (Grant No. JCYJ20210324120012033), and Overseas Cooperation Research Fund of Tsinghua Shenzhen International Graduate School  (Grant No. HW2021008).

\printcredits

\bibliographystyle{cas-model2-names}

\bibliography{cas-refs}

\bio{Figures/liyangning}
Yangning Li received the BEng degree from the Department of Computer Science and Technology, Huazhong University of Science and Technology, in 2020. He is currently working toward a Master's degree with the Tsinghua Shenzhen International Graduate School, Tsinghua University. His research interests include natural language processing and data mining.
\endbio
\vspace{1cm}

\bio{Figures/liyinghui}
Yinghui Li received the BEng degree from the Department of Computer Science and Technology, Tsinghua University, in 2020. He is currently working toward the PhD degree with the Tsinghua Shenzhen International Graduate School, Tsinghua University. His research interests include natural language processing and deep learning.
\endbio
\vspace{1cm}

\bio{Figures/chenxi}
Xi Chen received his PhD degree in computer science from the Zhejiang University. He is currently the head of the cross-modal algorithm center of Tencent Platform and Content Group and mainly focuses on various applications of NLP.
\endbio
\vspace{1cm}

\bio{Figures/zhenghaitao}
Hai-Tao Zheng received the bachelor’s and master’s degrees in computer science from the Sun Yat-Sen University, China, and the PhD degree in medical informatics from Seoul National University, South Korea. He is currently an associate professor with the Shenzhen International Graduate School, Tsinghua University, and also with Peng Cheng Laboratory. His research interests include web science, semantic web, information retrieval, and machine learning.
\endbio

\bio{Figures/shenying}
Ying Shen received the PhD degree in computer science from the University of Paris Ouest Nanterre La Défense, France and the Erasmus Mundus master's degree in natural language processing from the University of Franche-Comt, France and the University of Wolverhampton, U.K. She is currently an associate professor with the School of Intelligent Systems Engineering, Sun Yat-Sen University. Her research interests include natural language processing and deep learning.

\endbio
\vspace{1cm}
\end{document}